\documentclass{article}

\usepackage{microtype}
\usepackage{graphicx}
\usepackage{subfigure}
\usepackage{booktabs} 

\usepackage{hyperref}

\usepackage{epsfig,amsmath,amstext,psfrag,mathrsfs,amssymb,amsfonts,color,amsthm}
\usepackage{enumerate,cite}
\usepackage{graphicx,subfigure}
\usepackage{dsfont,xcolor}
\usepackage{indentfirst}

\usepackage[accepted]{icml2019}

\icmltitlerunning{Information-Theoretic Understanding of Population Risk Improvement with Model Compression}

\def\hW{{\hat{W}}}
\def\hw{{\hat{w}}}

\newtheorem{definition}{Definition}
\newtheorem{corollary}{Corollary}
\newtheorem{lemma}{Lemma}
\newtheorem{theorem}{Theorem}
\newtheorem{remark}{\textbf{Remark}}

\newcommand{\nn}{\nonumber}

\begin{document}

\twocolumn[
\icmltitle{Information-Theoretic Understanding of Population Risk Improvement with Model Compression}



\icmlsetsymbol{equal}{*}

\begin{icmlauthorlist}
\icmlauthor{Yuheng Bu}{uiuc}
\icmlauthor{Weihao Gao}{uiuc}
\icmlauthor{Shaofeng Zou}{buffalo}
\icmlauthor{Venugopal V. Veeravalli}{uiuc}
\end{icmlauthorlist}

\icmlaffiliation{uiuc}{ECE Department, University of Illinois at Urbana-Champaign, IL, USA}
\icmlaffiliation{buffalo}{Department of Electrical Engineering, University at Buffalo, the State University of New York, NY, USA}

\icmlcorrespondingauthor{Yuheng Bu}{bu3@illinois.edu}
\icmlcorrespondingauthor{Venugopal V. Veeravalli}{vvv@illinois.edu}

\icmlkeywords{ }

\vskip 0.3in
]



\printAffiliationsAndNotice{}  

\begin{abstract}

We show that model compression can improve the population risk of a pre-trained model, by studying the tradeoff between the decrease in the generalization error and the increase in the empirical risk with model compression. We first prove that model compression reduces an information-theoretic bound on the generalization error; this allows for an interpretation of model compression as a regularization technique to avoid overfitting. We then characterize the increase in empirical risk with model compression using rate distortion theory. These results imply that the population risk could be improved by model compression if the decrease in generalization error exceeds the increase in empirical risk. We show through a linear regression example that such a decrease in population risk due to model compression is indeed possible. Our theoretical results further suggest that the Hessian-weighted $K$-means clustering compression approach can be improved by regularizing the distance between the clustering centers. We provide experiments with neural networks to support our theoretical assertions.

\end{abstract}

\section{Introduction}
\label{sec:intro}
The recent success of deep neural networks has dramatically boosted the applications of machine learning \cite{krizhevsky2012imagenet,silver2017mastering,goodfellow2016deep}. However, implementing a deep neural network model  on resource-limited devices becomes increasingly difficult, as deep neural networks usually have a large number of parameters. For example, for the problem of  image classification, it takes over 200MB to save the parameters of AlexNet \cite{krizhevsky2012imagenet},  and more than 500MB to save the parameters of VGG-16 net \cite{simonyan2014very}. It is difficult to port such large models on to mobile devices and embedded systems, due to their limited storage, bandwidth, energy and computational resources.

For this reason there has been a flurry of recent work on compressing the coefficients of deep neural networks (see \cite{cheng2017survey,krishnamoorthi2018quantizing,guo2018survey} for recent surveys).
Existing studies mainly focus on designing compression algorithms to reduce the memory requirement and computational cost, while keeping the same population risk. However, in some recent works \cite{choi2016towards,zhu2016trained,lin2017deep}, it is observed empirically that the population risk of the compressed model can sometimes be \emph{better} than that of the original model.
This phenomenon is counterintuitive at a first glance, since compression generally leads to information loss. 

Indeed, as neural networks are usually trained by minimizing the empirical risk, a compressed model  has a larger empirical risk than the original one. Despite of this fact,  in this paper we show that model compression improves the generalization error, since it can be interpreted as a regularization technique to avoid overfitting. As the population
risk is the sum of the empirical risk and the generalization error, it is possible for the population risk to be reduced by model compression. 


\subsection{Contributions}


In this paper, we provide a comprehensive information-theoretic explanation for the population risk improvement with model compression by characterizing the tradeoff between the generalization error and the empirical risk. Specifically, we focus on the case where the model is compressed based on a pre-trained model.

\begin{figure}
  \centering
  \includegraphics[width=0.45\textwidth]{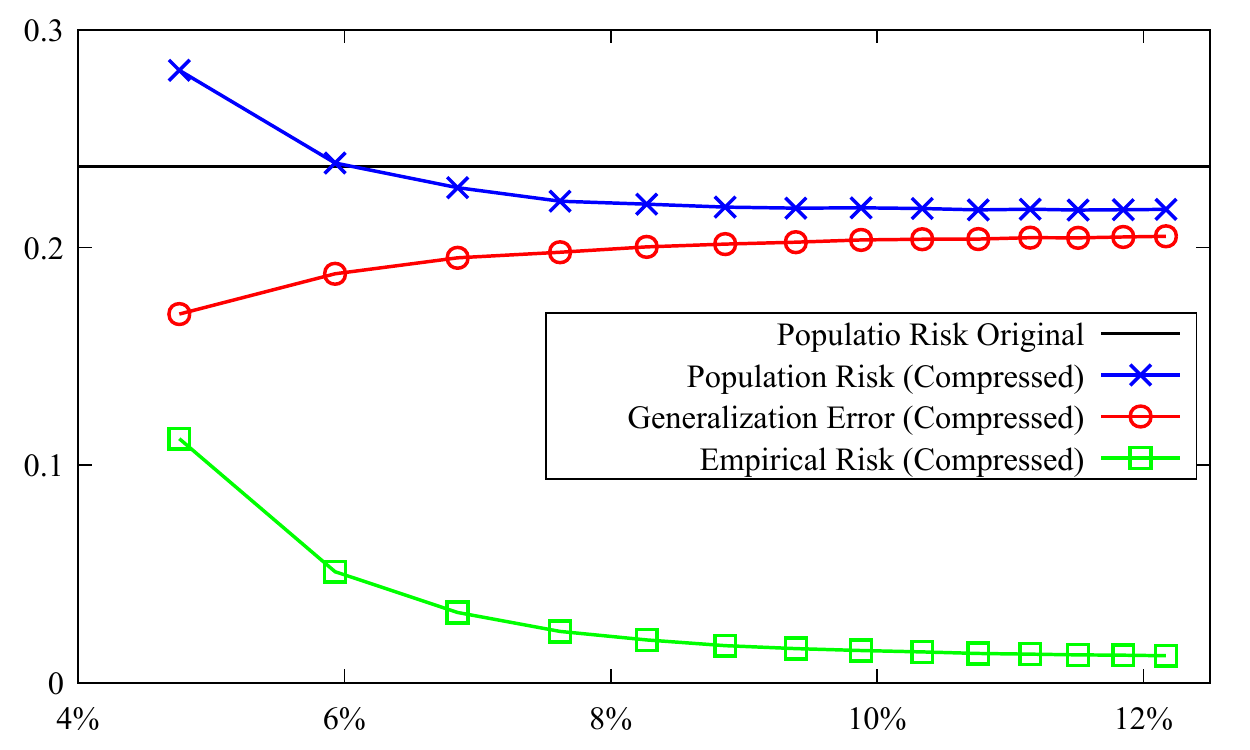}
  \put(-145,-10){Compression Ratio}
  \put(-225,25){\rotatebox{90}{Cross Entropy Loss}}
  \caption{Population risk of the compressed model $\hW$ and the original model $W$ vs.\ compression ratio. The generalization error of $\hW$ decreases and the empirical risk of $\hW$ increases with more compression, i.e., as the compression ratio gets smaller. The population risk of $\hW$ is less than that of $W$.}\label{Fig:illus}
  \vspace{-0.2cm}
\end{figure}

We first prove that model compression  tightens the information-theoretic generalization error bound in \cite{raginsky2016information}, and it can therefore be interpreted as a regularization method to reduce overfitting. Furthermore, we define the distortion as the difference in the empirical risk between the original and compressed models, and use rate distortion theory to characterize the distortion as a function of the number of bits $R$ used to describe the model.
We show if the decrease in generalization error exceeds the increase in empirical risk, the population risk can be improved.
An empirical illustration of this result for the MNIST dataset is provided in Figure \ref{Fig:illus}, where model compression and population risk improvement are achieved simultaneously (details are given in Section \ref{sec:exp}).
To better demonstrate our theoretical results, we investigate an example of linear regression comprehensively, where we develop explicit bounds on the generalization error and the distortion.





Our generalization error bound also suggests that the Hessian-weighted $K$-means clustering compression approach \cite{choi2016towards} can be improved by further regularizing the distance between the clustering centers. Our numerical experiments with neural networks validate our theoretical assertions and demonstrate the effectiveness of the proposed regularizer.


\subsection{Related Works}
There have been many recent studies on model compression for deep neural networks.  The compression could happen by varying the training process, e.g., network structure optimization  \cite{howard2017mobilenets}, low precision neural networks \cite{gupta2015deep} and   neural networks with binary weights \cite{courbariaux2015binaryconnect,rastegari2016xnor}. Here we mainly discuss  compression approaches that are applied on a pre-trained model, which does not require training a new neural network.

Pruning, quantization and matrix factorization are the most popular approaches to compress pre-trained deep neural networks. The study of pruning algorithms for model compression which remove redundant parameters from neural networks dates back to \cite{mozer1989skeletonization,lecun1990optimal,hassibi1993second}. More recently, \cite{han2015learning} propose an iterative pruning and retraining algorithm to further reduce the size of deep models. In addition, the method of network quantization or weight sharing is investigated in \cite{gong2014compressing,han2015deep,choi2016towards}, where a hard clustering algorithm is employed to group the weights in a neural network. Matrix factorization, i.e., low-rank approximation of the weights in neural networks has also been widely studied  \cite{denton2014exploiting,tai2015convolutional,novikov2015tensorizing}.

All of the aforementioned works demonstrate the effectiveness of their methods via comprehensive numerical experiments. Little research has been done to develop a theoretical understanding of how model compression affects performance. Some recent work includes \cite{gao2018rate} in which  an information-theoretic view of model compression via rate-distortion theory is provided, with the focus  on minimizing the empirical risk of the compressed model; and \cite{zhou2018non} in which a non-vacuous generalization error bound based on the small complexity of the compressed model is provided using a PAC-Bayesian framework.

In contrast to these works, the studies in this paper are from a different perspective, which is of more practical interest, i.e., the population risk of the compressed model. We develop an understanding as to why  model compression can  improve population risk based on an analysis of both the empirical risk and generalization error. More importantly, our theoretical studies offer insights on designing practical model compression algorithms. Specifically, the increase in empirical risk and the decrease in generalization error shall be considered jointly, so that the population risk can be improved.





\section{Preliminaries}\label{sec:model}

{\bf Notation:} We use upper letters to denote random variables, and calligraphic upper letters to denote sets. For a random variable $X$ generated from a distribution $\mu$, we use $\mathbb{E}_{X\sim\mu}$ to denote the expectation taken over $X$ with distribution $\mu$.  We use $I_d$ to denote the $d$-dimensional identity matrix, and $\|A\|$ to denote the spectral norm of a matrix $A$. The cumulant generating function (CGF) of a random variable $X$ is defined as $\Lambda_X(\lambda) \triangleq \ln \mathbb{E}[e^{\lambda(X-\mathbb{E}X)}]$. All logarithms are natural ones.

\subsection{Generalization Error}
Consider an instance space $\mathcal{Z}$, a hypothesis space $\mathcal{W}$, and a nonnegative loss function $\ell: \mathcal{W}\times \mathcal{Z} \to \mathbb{R}^+$. A training dataset $ S=\{Z_1,\cdots,Z_n\}$ consists of $n$ i.i.d samples $Z_i \in \mathcal{Z}$  drawn from an unknown distribution $\mu$. The goal of a supervised learning algorithm is to find an output hypothesis $w \in \mathcal{W}$ that minimizes the {population risk}:
\begin{equation}
  L_\mu(w) \triangleq \mathbb{E}_{Z\sim \mu}[\ell(w,Z)]. 
\end{equation}
\vspace{-0.1cm}
In practice, $\mu$ is unknown, and therefore $L_\mu(w)$ cannot be computed directly. Instead,  the {empirical risk} of $w$ on the training dataset $S$ is studied, which is defined as
\begin{equation}
  L_S(w)\triangleq \frac{1}{n} \sum_{i=1}^n \ell(w,Z_i).
\end{equation}
A learning algorithm can be characterized by a randomized mapping from the training data set $S$ to a hypothesis $W$ according to a conditional distribution $P_{W|S}$. The {generalization error} of a supervised learning algorithm is the expected difference between the population risk of the output hypothesis  and its empirical risk on the training dataset:
\begin{equation}
  \mathrm{gen}(\mu,P_{W|S}) \triangleq \mathbb{E}_{W, S}[L_\mu(W)-L_S(W)],
\end{equation}
where the expectation is taken over the joint distribution $P_{S,W} = P_S\otimes P_{W|S}$. 

\subsection{Review of Rate Distortion Theory}\label{sec:RD_theory}

Rate distortion theory, firstly introduced by \cite{shannon1959coding}, is a major branch of information theory which studies the fundamental limits of lossy data compression. It addresses  the minimal number of bits per symbol, as measured by the rate $R$, to transmit a random variable $W$ such that the receiver can reconstruct $W$ without exceeding a given distortion $D$.

Specifically, let $W^m = \{W_1,W_2,\cdots,W_m\}$ denote a sequence of  $m$ i.i.d. random variables $W_i \in \mathcal{W}$ generated from a source distribution $P_W$. An encoder $f_m : \mathcal{W}^m \to \{1,2,\cdots, M \}$ maps the message $W^m$ into  a codeword, and a decoder $g_m : \{1,2,\cdots, M \} \to \hat{\mathcal{W}}^m$ reconstructs the message by an estimate $\hat{W}^m$ from the codeword, where $\hat{\mathcal{W}} \subseteq \mathcal{W}$ denotes the range of $\hW$.  A distortion metric $d : \mathcal{W} \times \mathcal{W} \to  \mathbb{R}^+$ quantifies the difference between the original and reconstructed messages. The distortion between sequences $w^m $ and $\hw^m$  is defined to be
\begin{equation}
  d(w^m,\hw^m) \triangleq \frac{1}{m}\sum_{i=1}^m d(w_i,\hw_i).
\end{equation}
A commonly used distortion metric is the square distortion function: $d(w, \hat{w}) = (w-\hat{w})^2$, where $\mathcal{W} = \mathbb{R}$.
\begin{definition}
An $(m,M,D)$-code is achievable, if there exists a (probabilistic) encoder-decoder pair $(f_m, g_m)$ such that the alphabet of codeword has size $M$ and the expected distortion
$\mathbb E[d(W^m; g_m(f_m(W^m)))] \le D$.
%
\end{definition}
\begin{definition}\label{def:R_D}
The rate-distortion function $R(D)$ and the distortion-rate function $D(R)$ are defined as
\begin{align}
  R(D) &\triangleq \lim_{m\to \infty} \frac{1}{m} \log_2 M^*(m,D),\\
  D(R) &\triangleq \lim_{m\to \infty} D^*(m,R),
\end{align}
where $M^*(m,D)\triangleq\min \{M : (m,M,D) \text{ is achievable} \}$, and $D^*(m,R)\triangleq\min \{D : (m,2^{mR},D) \text{ is achievable}\}$.
\end{definition}

The main theorem of rate distortion theory is as follows.
\begin{lemma}\cite{cover2012elements} \label{lemma:rate_distortion}
For an i.i.d. source $W$ with distribution $P_W$ and distortion function $d(w,\hw)$, it follows that
\begin{align}
  R(D) &= \min_{P_{\hW|W}: \mathbb{E}[d(W,\hW)]\le D} I(W;\hW),\\
  D(R) &= \min_{P_{\hW|W}: I(W;\hW)\le R}  \mathbb{E}[d(W,\hW)],
\end{align}
where $I(W;\hW) \triangleq \mathbb{E}_{W,\hW}[\ln \frac{P_{W,\hW}}{P_W  P_\hW}]$ denotes the mutual information between $W$ and $\hW$.
\end{lemma}

The rate-distortion function quantifies the smallest number of bits required to compress the data given the distortion, and the distortion-rate function quantifies the minimal distortion that can be achieved under the rate constraint.



\section{Compression Improves Generalization}\label{sec:compress}
In this section, we prove that a lossy compression algorithm can be used to improve the generalization error of a supervised learning algorithm via an information-theoretic  generalization error bound.

\subsection{Information-theoretic Generalization Bounds}

The following lemma from \cite{raginsky2016information} provides an upper bound on the generalization error using the mutual information $I(S;W)$ between the training data set $S$ and the output of the learning algorithm $W$.
\begin{lemma}\cite{xu2017information}\label{lemma:original}
Suppose $\ell(w,{Z})$ is $\sigma$-sub-Gaussian \footnote{A random variable $X$ is $\sigma$-sub-Gaussian if $\Lambda_X(\lambda)\le \frac{\sigma^2\lambda^2}{2}$, $\forall \lambda \in \mathbb{R}$.
} under ${Z}\sim \mu$ for all $w \in \mathcal{W}$, then
\begin{equation}
  |\mathrm{gen}(\mu,P_{W|S}) | \le \sqrt{\frac{2\sigma^2}{n} I(S;W)}.
\end{equation}
\end{lemma}

We note that this mutual information based generalization error bound depends on the properties of the supervised learning problem, e.g., the hypothesis space $\mathcal{W}$, the learning algorithm $P_{W|S}$, the distribution of the training data $\mu$ and the loss function $\ell(w,Z)$.
%

\subsection{Information-theoretic Generalization Error Bound for Compressed Model}

Compression can be viewed as a post-processing of the output of a learning algorithm. The output model $W$ generated by a learning algorithm can be quantized, pruned or even perturbed by noise, which results in a compressed model $\hW$. Assume that the compression algorithm is only based on $W$, and can be described by a conditional distribution  $P_{\hW|W}$. Then the following Markov chain holds:  $S\to W \to \hW$, and by the data processing inequality,
\begin{equation}
I(S;\hW)\le \min\{I(W;\hW), I(S,W)\}.
\end{equation}
Thus, we have the following theorem that characterizes the generalization error of the compressed model $\hW$. 

\begin{theorem}\label{thm:compression}
Consider a learning algorithm $P_{W|S}$, a compression algorithm $P_{\hW|W}$, and suppose $\ell(\hw,{Z})$ is $\sigma$-sub-Gaussian under ${Z}\sim \mu$ for all $\hw \in \hat{\mathcal{W}}$. Then
\begin{equation*}
  |\mathrm{gen}(\mu,P_{\hW|S}) |  \le  \sqrt{\frac{2\sigma^2}{n} \min\{I(W;\hW), I(S,W)\}}.
\end{equation*}
\end{theorem}
Note that the generalization error bound in Theorem \ref{thm:compression} for the compressed model is tighter  than the one in Lemma \ref{lemma:original}.
Thus, a compression algorithm can be interpreted as a regularization technique to reduce the generalization error.

\section{Tradeoff between Generalization Error and Distortion}\label{sec:distortion}
In this section, we first define the distortion metric in model compression, and then connect the distortion with the  generalization error bound using rate-distortion theory. We show that there is a tradeoff between the generalization error and the distortion, and we can improve the population risk if the decrease in generalization error exceeds the increase in empirical risk.

\subsection{Distortion Metric in Model Compression}
The expected population risk of  $W$ can be written as
\begin{equation}\label{eq:Test_loss}
   \mathbb{E}_{S,W}[L_\mu(W)] = \mathbb{E}[L_S(W)] + \mathrm{gen}(\mu,P_{W|S}) ,
\end{equation}
where the first term, which is the expected empirical risk, reflects how well the model $W$ fits the training data, while the second term demonstrates how well the model generalizes. In the empirical risk minimization (ERM) framework, we minimize both terms by 1) minimizing the empirical risk of $W$ directly or using other stochastic optimization algorithms, and 2) using regularization methods to control the generalization error, e.g., early stopping and dropout  \cite{goodfellow2016deep}.


Consider the expected population risk of the compressed model $\hW$,
\begin{align}
  & \mathbb{E}_{S,W,\hat{W}}[ L_\mu(\hW))] \nn \\
  &= \mathbb{E}[ L_\mu(\hW)- L_S(\hW) + L_S(\hW) -L_S(W) + L_S(W)] \nn \\
  &= \mathbb{E}[L_S(W)] + \mathrm{gen}(\mu,P_{\hW|S}) + \mathbb{E}[L_S(\hW) -L_S(W)].\nn
\end{align}
Compared with \eqref{eq:Test_loss}, we note that the first empirical risk term is independent of the compression algorithm, the second generalization error term can be upper bounded by Theorem \ref{thm:compression}, and the third term $\mathbb{E}[L_S(\hW) -L_S(W)]$ quantifies the distortion in the empirical risk if we use the compressed model $\hW$ instead of the original model $W$. We then define the following distortion metric for model compression:
\begin{equation}\label{eq:distortion_model}
d_S(w, \hw) \triangleq L_S(\hw) -L_S(w),
\end{equation}
which is the difference in the empirical risk between the compressed model $\hW$ and the original model $W$.
By Theorem \ref{thm:compression}, it follows that
\begin{align}\label{eq:test_upperbound}
  &\mathbb{E}_{S,W,\hat{W}}[ L_\mu(\hW)-L_S(W)]  \nn \\
  & \le \sqrt{\frac{2\sigma^2}{n}I(W;\hW)} + \mathbb{E}_{S,W,\hat{W}}[d_S (\hW,W)]\nn \\
  &\triangleq \mathcal{L}_{S,W}(P_{\hW|W}),
\end{align}
where $\mathcal{L}_{S,W}(P_{\hW|W})$ is an upper bound on the expected difference between the population risk of $\hW$ and the empirical risk of the original model $W$ on training dataset $S$.

\subsection{Population Risk Improvement}
Suppose that we use $R$ bits to quantize $W$, i.e., $I(W;\hat W)=R$. By  Lemma \ref{lemma:rate_distortion}, the smallest distortion that can be achieved at rate $R$ is
\begin{equation}
 D(R) = \min_{I(W;\hW) \le R} \mathbb{E}_{S,W,\hat{W}}[d_S (\hW,W)].
\end{equation}
In this case, the tightest bound in \eqref{eq:test_upperbound} that can be achieved at rate $R$  is given in the following theorem.
\begin{theorem}\label{thm:tradeoff}
	Suppose the assumptions in Theorem \ref{thm:compression} hold, and $I(W;\hat W)=R$, then
	\begin{align}
	\min_{P_{\hW|W}:I(W;\hW) = R} &\mathbb{E}_{S,W,\hat{W}}[ L_\mu(\hW)-L_S(W)]\nn\\
	&\leq \sqrt{\frac{2\sigma^2 }{n}R }+D(R).
	\end{align}
\end{theorem}

%
From the properties of the distortion-rate function \cite{cover2012elements}, we know that $D(R)$  is a decreasing function of $R$. Thus, to minimize the population risk of the compressed model $\hW$,
%
there is a tradeoff between the rate $R$, which upper bounds the generalization error, and the distortion $D(R)$
on the empirical risk. Such a tradeoff is similar to the relationship between the complexity of the hypothesis space, e.g., VC dimension, and the empirical risk, where a simple and small model could have a small generalization error, but may underfit the training data.
%
Theorem \ref{thm:tradeoff} further suggests that we can possibly improve the population risk of $\hW$ if we can make the decrease in the generalization error greater than the increase in the distortion of the empirical risk. As will be shown Section \ref{sec:exp}, such a tradeoff can be observed in practice, and it is possible to improve the population risk of $\hW$ with a properly chosen compression algorithm and compression ratio.



\section{Example: Linear Regression} \label{sec:linear}
In this section, we comprehensively explore the example of linear regression to get a better understanding of the results in Section \ref{sec:distortion}. To this end, we develop explicit upper bound for generalization error and distortion-rate function $D(R)$. All the proofs are provided in the supplementaries.

Suppose that the dataset $S=\{Z_1,\cdots,Z_n\}=\{(X_1,Y_1),\cdots,(X_n,Y_n)\}$ is generated from the following linear model with weight vector $w^*=(w^{*(1)},\cdots,w^{*(d)}) \in \mathbb{R}^{d}$,
\begin{equation}\label{eq:linear_model}
Y_i=X_i^\top w^*+\varepsilon_i,\, i=1,\cdots,n,
\end{equation}
where $X_i$'s are i.i.d. $d$-dimensional random vectors with distribution $\mathcal{N}(0, \Sigma_X)$, and $\varepsilon_i \sim  \mathcal{N}(0,\sigma'^2)$ denotes i.i.d. Gaussian noise. 
We adopt the mean squared error as the loss function, and the empirical risk on $S$ is
\begin{equation}
  L_S(w)=\frac{1}{n} \sum_{i=1}^n (Y_i-X_i^\top w)^2 = \frac{1}{n} \|Y-X^\top w\|_2^2,
\end{equation}
for $w\in \mathcal{W}=\mathbb{R}^d$, where $X\in \mathbb{R}^{d \times n}$ denotes all the input samples, and $Y \in \mathbb{R}^n$ denotes the responses. If $n>d$, the ERM solution is
\begin{equation}\label{eq:linear_ERM}
  W = (X X^\top)^{-1} X Y,
\end{equation}
which is deterministic given $S$. Its generalization error can be computed exactly as in the following lemma.

\begin{lemma}\label{lemma:linear_gen}
If $n>d+1$, then
\begin{equation}
\mathrm{gen}(\mu,P_{W|S}) = \frac{\sigma'^2d}{n}(2+\frac{d+1}{n-d-1}).
\end{equation}
\end{lemma}



\subsection{Information-theoretic Generalization Bounds for Compressed Linear Model}
We note that the mutual information based bound in Lemma \ref{lemma:original} is not applicable for this linear regression model, since $W$ is a deterministic function of $S$, and $I(S;W)=\infty$. However, this issue can be solved if we post-process the ERM solution $W$ by a compression algorithm, and use Theorem \ref{thm:compression} to upper bound the generalization error by $I(\hW;W)$.



Consider a compression algorithm, which maps the original weights $W \in \mathbb{R}^d$ to the compressed model $\hW\in \hat{\mathcal{W}} \subseteq \mathbb{R}^d$. For a fixed and compact $\hat{\mathcal{W}}$, we define
\begin{equation}
 C(w^*) \triangleq \sup_{\hw \in \hat{\mathcal{W}}}\|\hw-w^*\|_2^2,
\end{equation}
which measures the largest distance between the reconstruction $\hw$ and the optimal weights $w^*$.

The following theorem provides an upper bound on the generalization error of the compressed model $\hW$. 
\begin{theorem}\label{thm:regression}
Consider the ERM solution defined in \eqref{eq:linear_ERM}, and suppose $\hat{\mathcal{W}}$ is compact, then
\begin{equation}
  \mathrm{gen}(\mu,P_{\hW|S})  \le 2\sigma_\ell^{*2} \sqrt{\frac{I(W;\hW)}{n} }.
\end{equation}
where $\sigma_\ell^{*2} \triangleq C(w^*) \|\Sigma_X\|+\sigma'^2$.
\end{theorem}

\subsection{Distortion-Rate Function for Linear Model}
In this subsection, we provide an upper bound on the distortion-rate function $D(R)$ for linear regression model, and further demonstrate the tradeoff between generalization error and distortion.

Note that $\nabla L_S(W)=0$, since $W$ minimizes the empirical risk. The Hessian matrix of the loss function is
\begin{equation}
H_S(W) = \frac{1}{n} X X^\top,
\end{equation}
which is not a function of $W$. Then distortion function can be written as follows:
\begin{align}\label{eq:linear_distortion}
  &\mathbb{E}_{S,W,\hat{W}}[d_S (\hW,W)] \nn \\
  & = \mathbb{E}_{S,W,\hat{W}}[ L_S(\hW) -L_S(W)] \nn \\
   & = \mathbb{E}_{S,W,\hat{W}}[(\hW-W)^\top \frac{1}{n} X X^\top (\hW-W)].
 \end{align}
The following theorem characterizes upper bounds for $R(D)$ and $D(R)$ for linear regression.
\begin{theorem}\label{thm:DR}
For the ERM solution $W$  in \eqref{eq:linear_ERM}, we have
\begin{align}
  & R(D) \le  \frac{d}{2} \Big(\ln  \frac{d\sigma'^2}{(n-d-1)D} \Big)^+, \quad D\ge 0, \\
  & D(R) \le  \frac{d\sigma'^2}{n-d-1}e^{-\frac{2R}{d}},\quad R\ge 0,
\end{align}
where $(x)^+=\max\{0,  x\}$.
\end{theorem}
\begin{remark}\label{remark:RD}
As shown in \cite{vershynin2010introduction}, if $n = O(d/\epsilon^2)$, $\|\frac{1}{n} X X^\top - \Sigma_X\| \le \epsilon$ holds with high probability. Then, the following lower bound on $R(D)$ holds if we can approximate $\frac{1}{n} X X^\top$ in \eqref{eq:linear_distortion} using $\Sigma_X$,
\begin{equation}
  R(D) \ge \frac{d}{2} \Big(\ln \frac{d\sigma'^2}{(n-d-1)D}\Big)^+ -D(P_W\|P_{W_G}),
\end{equation}
where $W_G$ denotes a Gaussian random vector with the same mean and variance as $W$.
\end{remark}
The proof of the upper bound for $R(D)$ is based on considering a Gaussian random vector 
which has the same mean and covariance matrix as $W$. In addition, the upper bound is achieved when $W-\hat{W}$ is independent of the dataset $S$ with the following conditional distribution,
\begin{equation}\label{eq:achievability}
 P_{\hW|W} = \mathcal{N}\big((1-\alpha)W+ \alpha w^*, (1-\alpha) \frac{D}{d}\Sigma_X^{-1}\big),
\end{equation}
where $\alpha \triangleq \frac{nD}{d\sigma'^2} \le 1$. Note that this ``compression algorithm'' requires the knowledge of optimal weights $w^*$, which is unknown in practice. 


Combing Theorems \ref{thm:regression} and  \ref{thm:DR}, we have the following result.
\begin{corollary}\label{cor:tradeoff_linear}
Under the same assumptions as in Theorem \ref{thm:regression}, we have
\begin{flalign}\label{eq:tradeoff_linear}
&\min_{P_{\hW|W}:I(\hat{W};\hW) = R} \mathbb{E}_{S,W,\hat{W}}[ L_\mu(\hW)-L_S(W)]\nn\\
&\qquad \le  2\sigma_\ell^{*2} \sqrt{\frac{R}{n}} + \frac{d\sigma'^2}{n-d-1}e^{-\frac{2R}{d}},\quad R\ge 0.
\end{flalign}
\end{corollary}

It is clear that in \eqref{eq:tradeoff_linear} the first term corresponds to the generalization error, which decreases with compression, and the second term corresponds to the empirical risk, which increases with compression. Thus, if the first term decreases more than the second term increases, we can obtain a smaller population risk.

\subsection{Evaluation and Visualization}

In the following plots, we generate the training data set $S$ using the linear model in \eqref{eq:linear_model} by letting $d=50$, $n=80$, $\Sigma_X=I_d$ and $\sigma'^2=1$.
We consider the following two compression algorithms. The first one is the conditional distribution $P_{\hW|W}$ in the proof of achievability \eqref{eq:achievability}, which requires the knowledge of the optimal weights $w^*$ and is denoted as ``Oracle''. The second is the well-known $K$-means clustering algorithm, where the weights in $W$ are grouped into $K$ clusters and represented by the cluster centers in the reconstruction $\hW$. By changing the number of clusters $K$, we can control the rate $R $, i.e., $ I(W;\hW)$.

\begin{figure}[t!]
\begin{minipage}[b]{0.48\textwidth}
\begin{center}
\centerline{\includegraphics[width=\columnwidth]{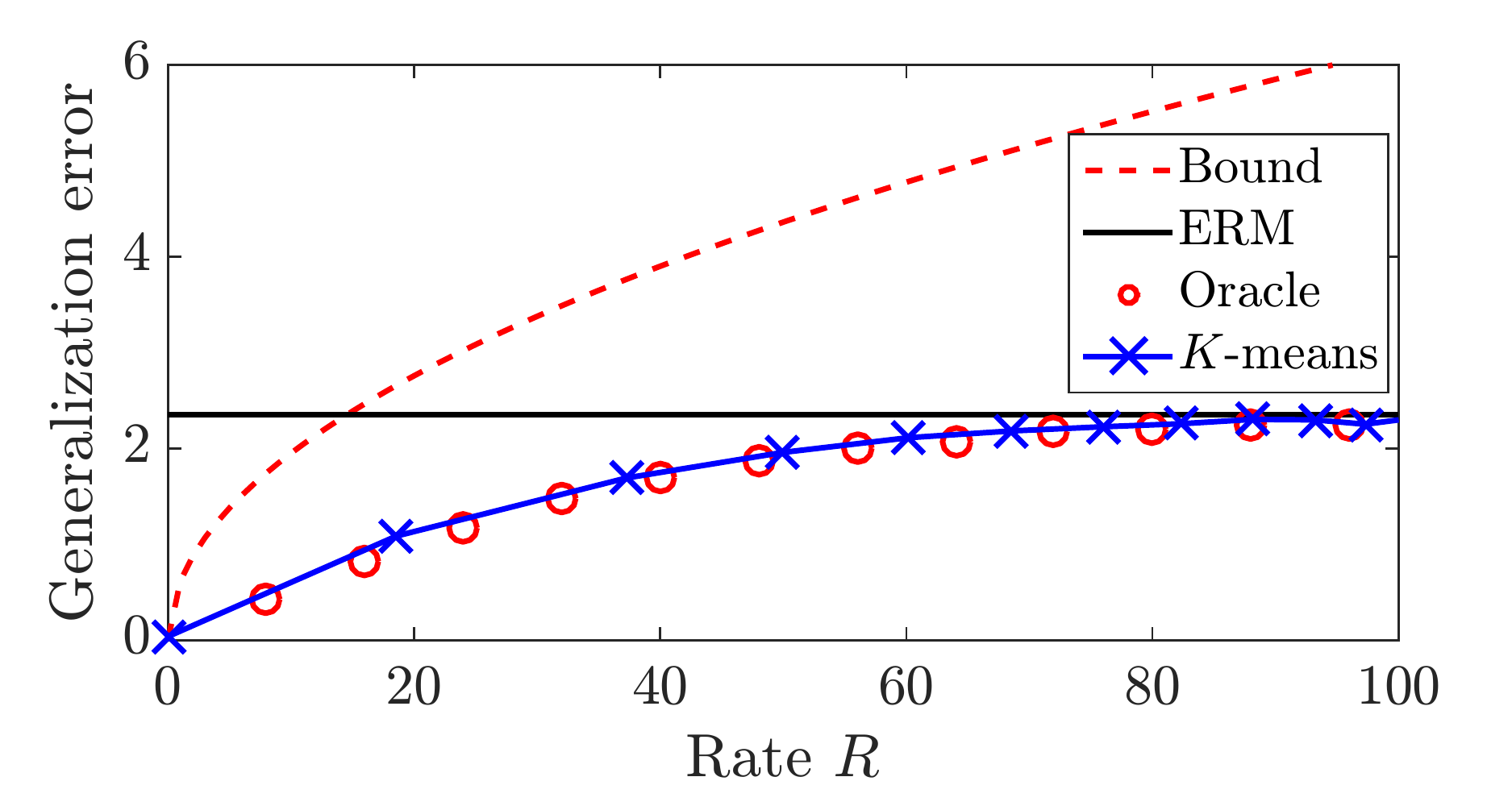}}
\caption{Comparison between the generalization error bound and generalization errors of different algorithms for linear regression.}
\label{Fig:gen_R}
\end{center}
\end{minipage}
\begin{minipage}[b]{0.48\textwidth}
\begin{center}
\centerline{\includegraphics[width=\columnwidth]{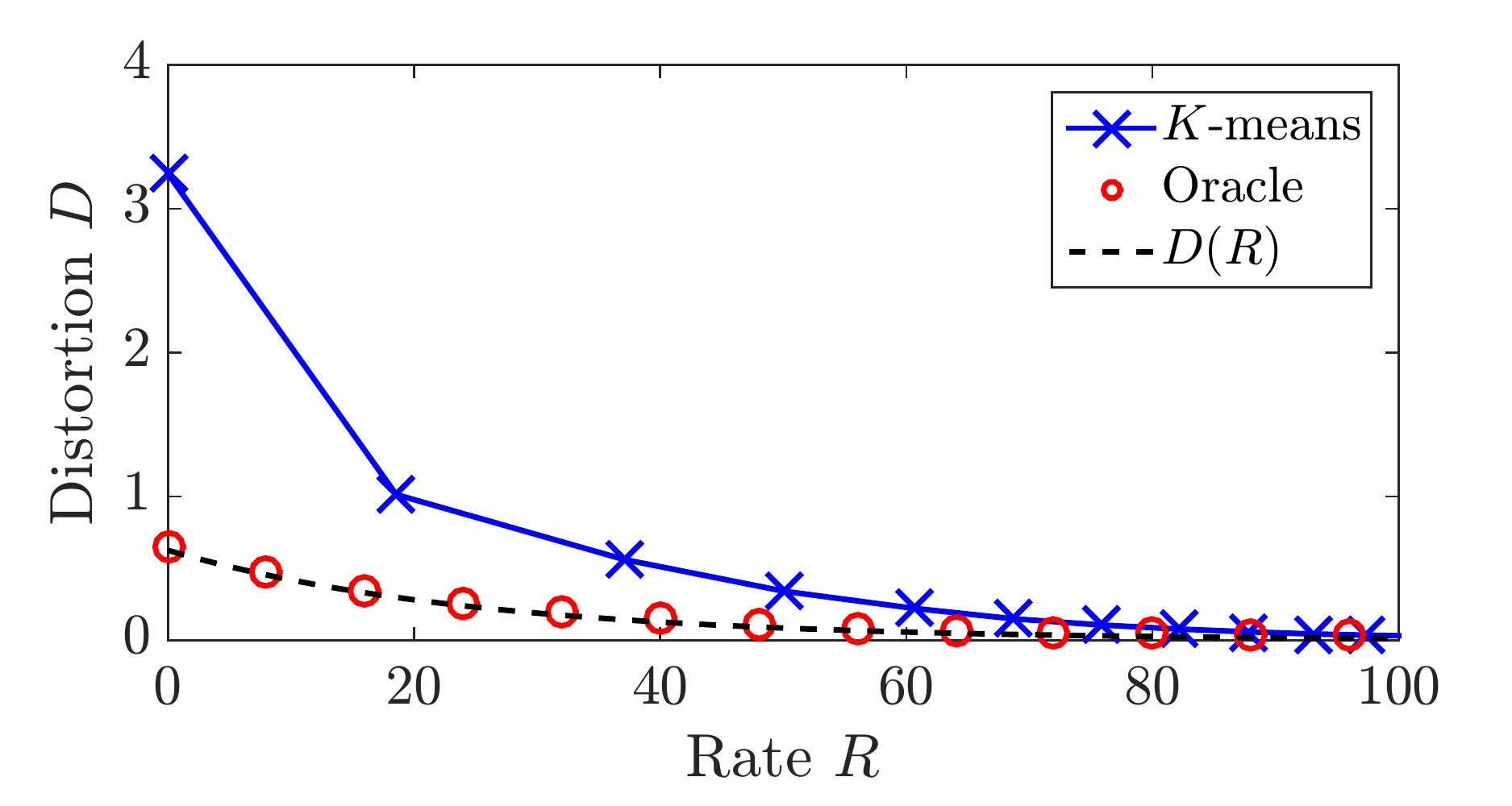}}
\caption{Distortions achieved by different algorithms for linear regression.}
\label{Fig:D_R}
\end{center}
\end{minipage}
\begin{minipage}[b]{0.48\textwidth}
\begin{center}
\centerline{\includegraphics[width=\columnwidth]{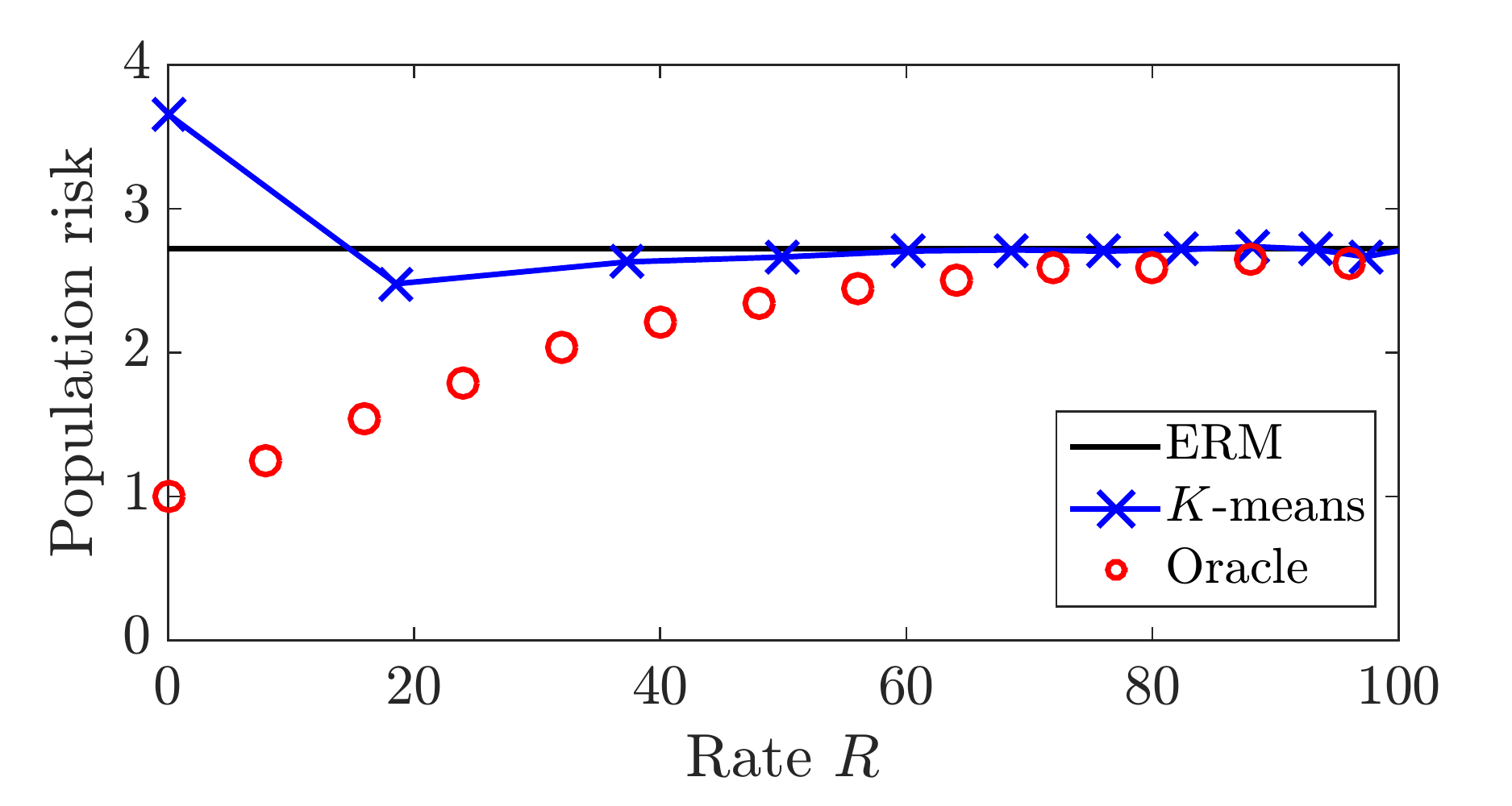}}
\caption{Comparison of the population risks achieved by different algorithms for linear regression.}
\label{Fig:test_R}
\end{center}
\end{minipage}
\vspace{-0.5cm}
\end{figure}
In Figure \ref{Fig:gen_R}, we plot the generalization error bound in Theorem \ref{thm:regression} as a function of the rate $R$, and compare the generalization errors of the Oracle and $K$-means algorithms. It can be seen that Theorem \ref{thm:regression} provides a valid upper bound for the generalization error, but this bound is tight only when $R$ is small. Moreover, both compression algorithms can achieve smaller generalization errors compared to that of the ERM solution $W$, which validates the argument in Theorem \ref{thm:compression}.

Figure \ref{Fig:D_R} plots the upper bound on the distortion-rate function in Theorem \ref{thm:DR} and the distortions achieved by the Oracle and $K$-means algorithms. The distortion of the Oracle algorithm decreases as we increase the number of bits to describe the model, and matches the $D(R)$ function well. However, there is a large gap between the distortion achieved by $K$-means algorithms and $D(R)$. One possible explanation is that since $w^*$ is unknown, it is impossible for the $K$-means algorithm to learn the optimal cluster center with only one sample of $W$. Even if we view $W^{(j)}, j=1,\cdots,d$ as i.i.d. samples from the same distribution, there is still a gap between the distortion achieved by the optimal quantization as studied in \cite{linder1994rates}. 

We plot the population risks of the ERM solution $W$, the Oracle and $K$-means algorithms in Figure \ref{Fig:test_R}. It is not surprising that the Oracle algorithm achieves a small population risk, since $\hW$ is a function of $w^*$ and $\hW=w^*$ when $R=0$. However, it can be seen that $K$-means algorithm achieves a smaller population risk than the original model $W$ by have the decrease in generalization error exceed the increase in empirical risk, when we use fewer clusters in the $K$-means algorithm. We note that the minimal population risk is achieved when $K=2$, since we initialize $w^*$ so that $w^{*(i)}$, $1\leq i\leq d$, can be well approximated by two clustering centers.

\section{Quantization Algorithm Minimizing $\mathcal{L}_{S,W}$} \label{sec:algorithms}
In this section, 
we show that the Hessian-weighted $K$-means clustering compression approach \cite{choi2016towards} can be improved by regularizing the distance between the clustering centers, which minimizes the upper bound $\mathcal{L}_{S,W}(P_{\hW|W})$, as suggested by our theoretical results.


\subsection{Hessian-weighted $K$-means Clustering}
A popular learning algorithm is ERM with stochastic optimization. Thus, the ERM solution $W$ is a local minimum of $L_S(w)$ and $\nabla L_S(W) = 0$, and therefore the distortion metric can be approximately written as
\begin{equation}
  d_S (\hW,W) \approx \frac{1}{2} (\hW-W)^\top H_S(W)(\hW-W),
\end{equation}
where $H_S(W)$ denotes the Hessian matrix of $L_S$ at $W$, and we ignore the higher order terms in the Taylor series expansion. For ease of  implementation, we further approximate the Hessian matrix by a diagonal matrix, i.e.,
\begin{equation}\label{eq:Hessian_distortion}
  d_S (\hW,W) \approx \frac{1}{2} \sum_{j=1}^d h^{(j)}(W^{(j)}-\hW^{(j)})^2,
\end{equation}
where $h^{(j)}$ is the $j$-th diagonal element of the Hessian matrix $H_S(W)$.
Thus, for a quantization algorithm, where the $d$-dimension parameters are partitioned into $K$ clusters, i.e., the size of codebook $M=K$, we can use the Hessian-weighted $K$-means clustering in \cite{choi2016towards} to minimize the distortion function in \eqref{eq:Hessian_distortion}.

Given network parameters $w=\{w^{(1)},\cdots, w^{(d)}\}$, the Hessian weighted $K$-means clustering partitions them into $K$ disjoint clusters, using a set of cluster centers $c=\{c^{(1)}, \cdots, c^{(K)}\}$, and a cluster assignment $C=\left\{C^{(1)}, \cdots, C^{(K)}\right\}$, while minimizing:
\begin{align}
    \min &\sum_{k=1}^K \sum_{w^{(j)} \in C^{(k)}}  h^{(j)}|w^{(j)}-c^{(k)}|^2.
\end{align}
\subsection{Diameter Regularization}

The goal of Hessian-weighted $K$-means clustering is to minimize the distortion on the empirical risk $L_S(\hW)$. However, a more interesting goal is to obtain as small a population risk as possible. This matches with the goal of our theoretical results, which is to reduce the population risk $L_{\mu}(\hW)$ by minimizing its upper bound  in \eqref{eq:test_upperbound}:
\begin{align}\label{eq:Reg_Loss}
\hspace{-0.1cm}\mathcal{L}_{S,W}(P_{\hW|W}) = \sqrt{\frac{2\sigma^2}{n}I(W;\hW)} + \mathbb{E}[d_S (\hW,W)].
\end{align}
In practice, we fix the codebook size $M=K$, then $I(W;\hW) \le \log_2 K$, and we want to minimize $\mathcal{L}_{S,W}(P_{\hW|W})$ by carefully designing the codebook, i.e., choosing $\{c^{(1)}, \cdots, c^{(K)}\}$.
For a fixed $n$, the sub-Gaussian parameter $\sigma$ is one way to control the generalization error of a compression algorithm. Recall that in Theorem \ref{thm:regression},
\begin{equation*}
  \mathrm{gen}(\mu,P_{\hW|S})  \le 2 \big(C(w^*) \|\Sigma_X\|+\sigma'^2\big)\sqrt{\frac{I(W;\hW)}{n} },
\end{equation*}
where the sub-Gaussian parameter is related to
$C(w^*) = \sup_{\hw \in \hat{\mathcal{W}}}\|\hw-w^*\|_2^2$ in  linear regression. Note that this quantity can be interpreted as the diameter of the codebook $\mathrm{Diam}(\hat{\mathcal{W}})$. Since the ground truth $w^*$ is unknown in practice, we then propose the following diameter regularization by approximating the first term in \eqref{eq:Reg_Loss} by
\begin{equation}
   \beta \max_{k_1,k_2}|c^{(k_1)}-c^{(k_2)}|^2, \beta \ge 0,
\end{equation}
where $\beta$ is a parameter controls the penalty term, and can be selected by cross validation in practice.

Our diameter-regularized Hessian-weighted $K$-means algorithm solves the following optimization problem:
\begin{equation*}
    \min \sum_{k=1}^K \sum_{w^{(j)} \in C^{(k)}}  h^{(j)}|w^{(j)}-c^{(k)}|^2 + \beta \max_{k_1,k_2}|c^{(k_1)}-c^{(k_2)}|^2.
\end{equation*}
An iterative algorithm to solve this optimization is provided in Appendix \ref{app:algorithm}.

\begin{figure}[tb!]
    \centering
    \includegraphics[width=0.45\textwidth]{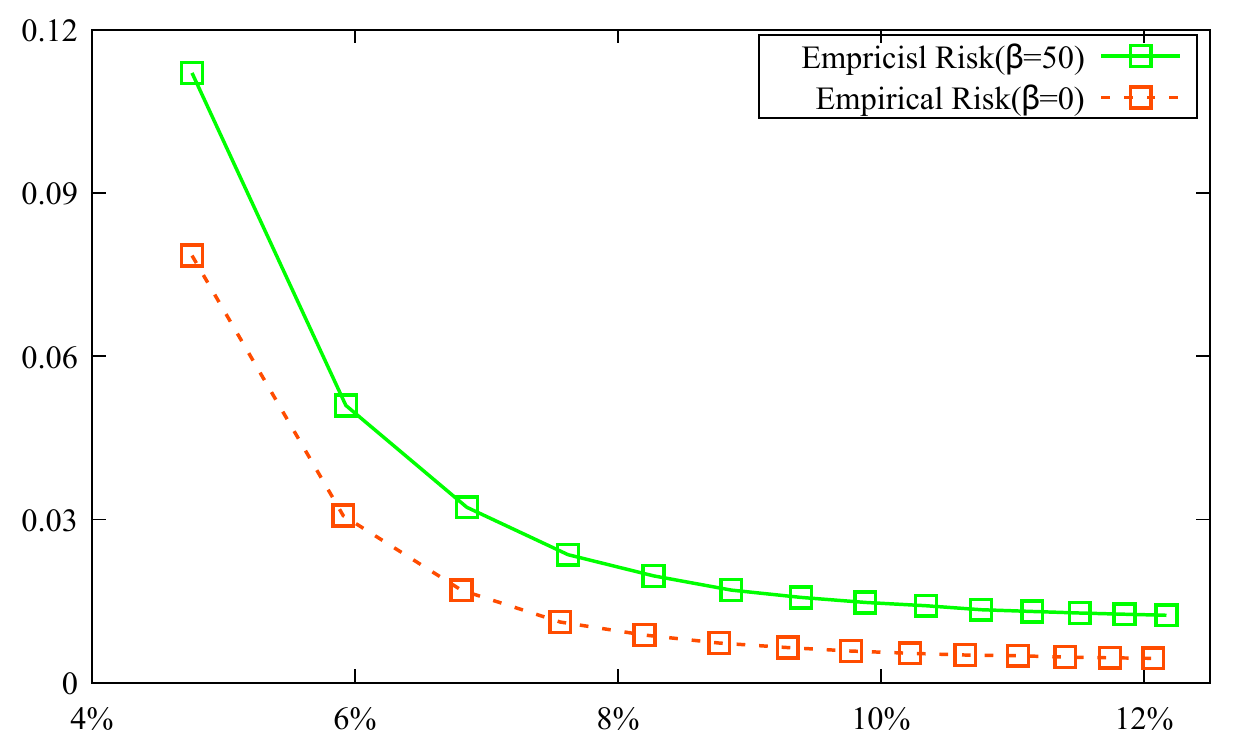}
    \put(-225,25){\rotatebox{90}{Cross Entropy Loss}} \\
    \includegraphics[width=0.45\textwidth]{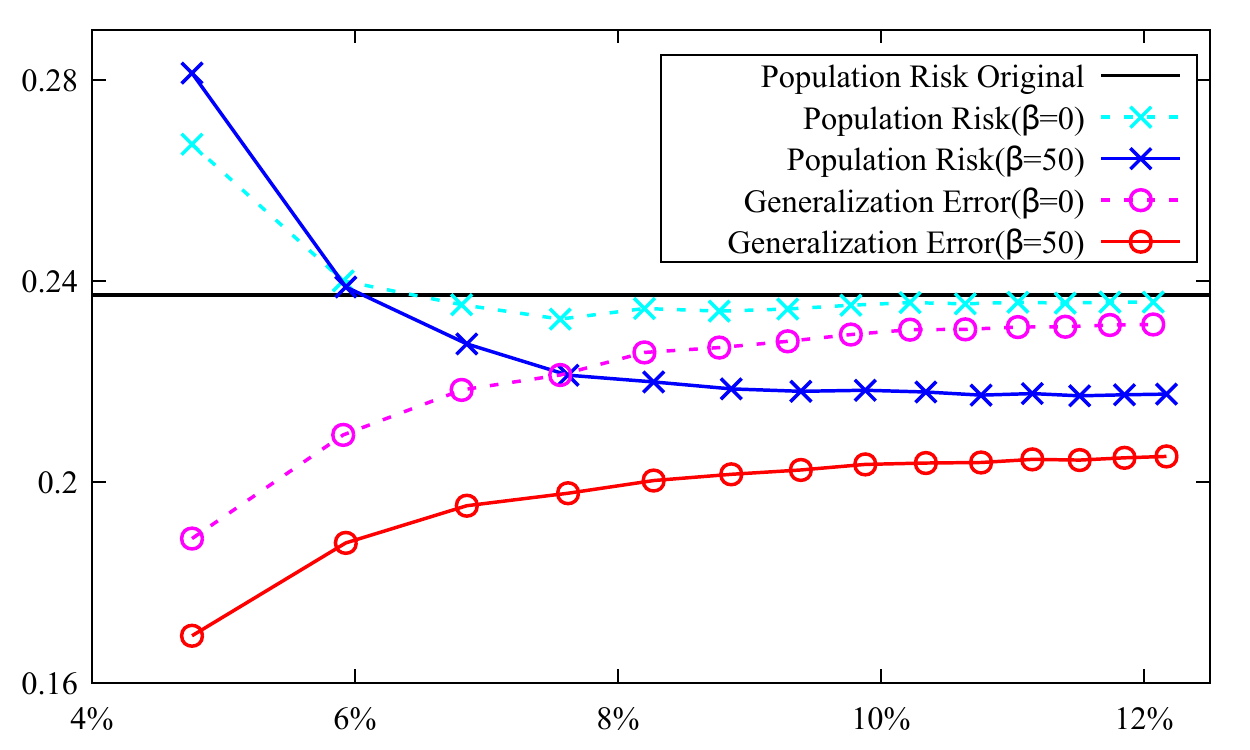}
    \put(-225,25){\rotatebox{90}{Cross Entropy Loss}}
    \put(-150,-10){Compression Ratio}
    \caption{Comparison between the diameter regularized Hessian weighted $K$-means algorithm ($\beta=50$) and the original one ($\beta=0$) on MNIST. Top: comparison of empirical risks. Bottom: comparison of population risks and generalization errors.}
    \label{Fig:mnist}
\end{figure}

\begin{figure}[tb!]
    \centering
    \includegraphics[width=0.45\textwidth]{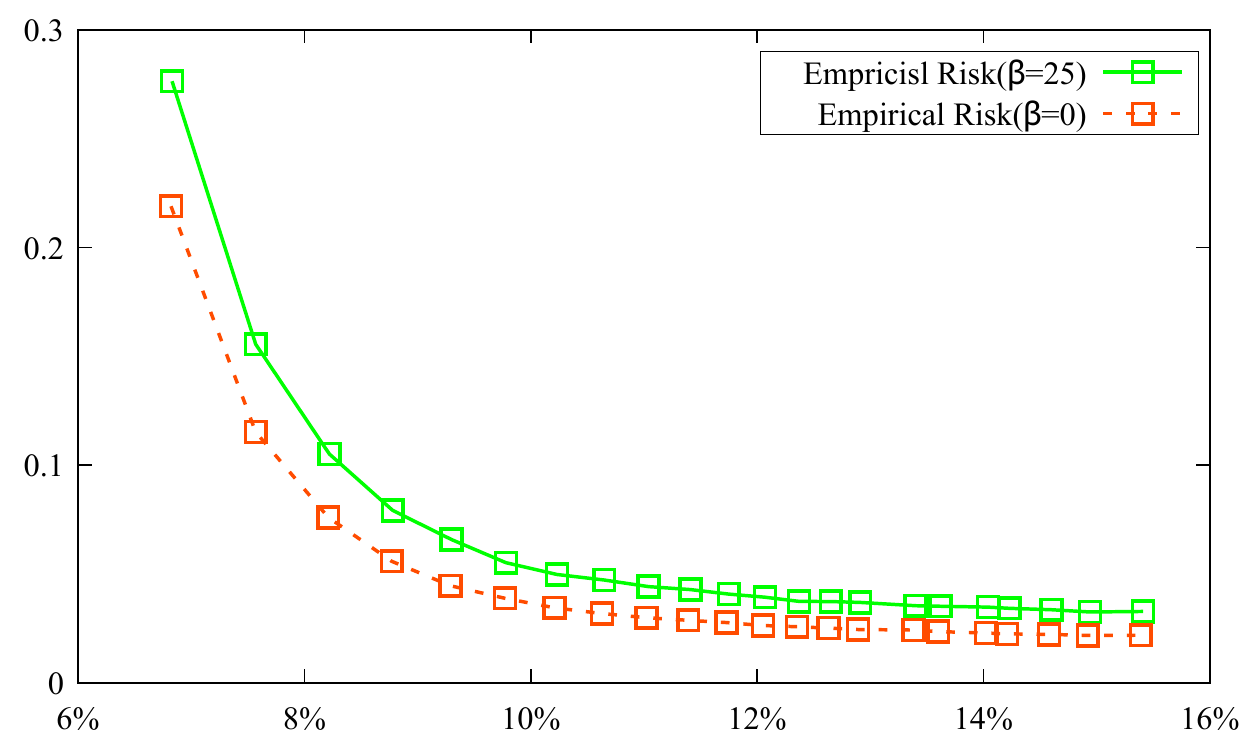}
    \put(-225,25){\rotatebox{90}{Cross Entropy Loss}} \\
    \includegraphics[width=0.45\textwidth]{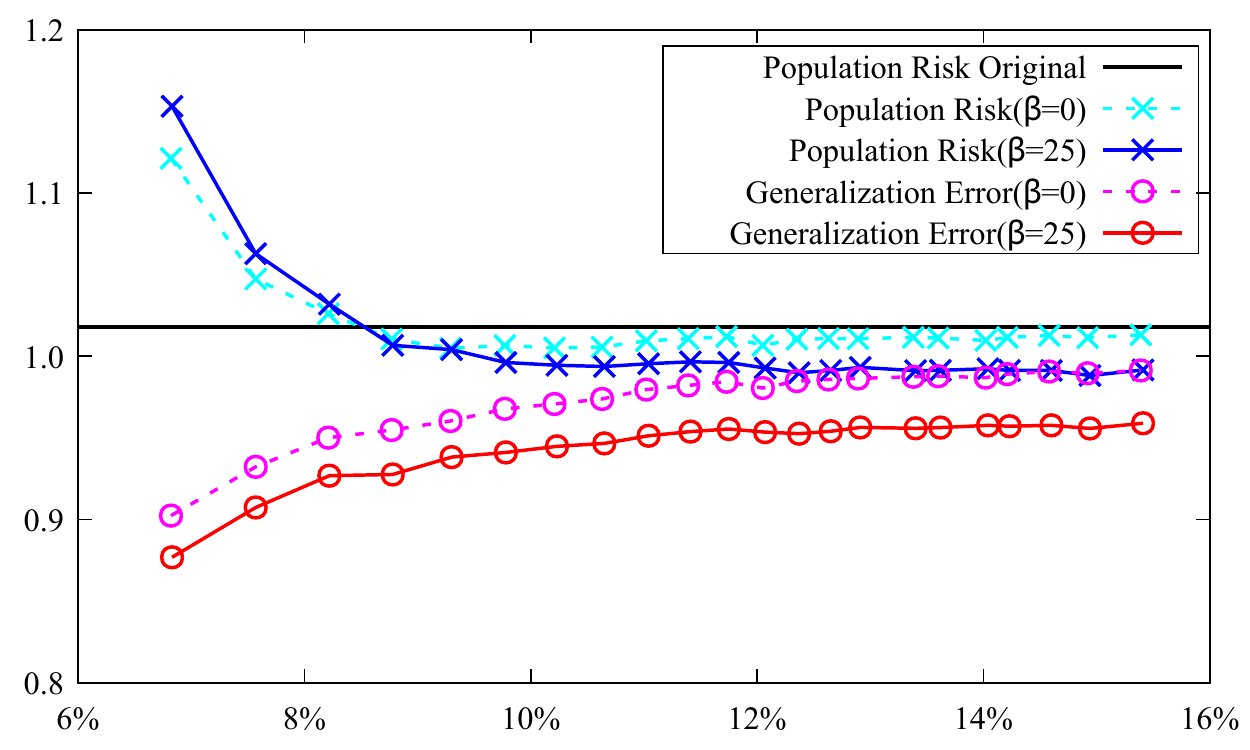}
    \put(-225,25){\rotatebox{90}{Cross Entropy Loss}}
    \put(-150,-10){Compression Ratio}
    \caption{Comparison between the diameter regularized Hessian weighted $K$-means algorithm ($\beta=25$) and the original one ($\beta=0$) on CIFAR10. Top: comparison of empirical risks. Bottom: comparison of population risks and generalization errors.}
    \label{Fig:CIFAR}
\end{figure}

\section{Experiments}\label{sec:exp}
In this section, we provide some real-world experiments to validate our theoretical assertions and the proposed diameter-regularized Hessian-weighted $K$-means algorithm. Our experiments include: (i) a 3-layer fully connected network on MNIST; and (ii) a convolutional neural network with 5 conv layers and 3 linear layers on CIFAR10 \footnote{The details of the network can be found in https://github.com/aaron-xichen/pytorch-playground.}.

In Theorem \ref{thm:compression}, an upper bound on the \emph{expected} generalization error is provided, so we independently train 50 different models (with the same structure but different parameter initializations), and average the results. We use 10\% of the training data to train the model for MNIST, and use 20\% of the training data to train the model for CIFAR10.
For each experiment, we use the same number of clusters for each convolutional layer and fully connected layer.

\subsection{Results}
In Figures \ref{Fig:mnist} and \ref{Fig:CIFAR}, we compare  our diameter-regularized Hessian-weight $K$-means algorithm and the original one with different compression ratios on the MNIST and CIFAR10 datasets. Both figures demonstrate that the proposed quantization algorithm increases the empirical risk, but decreases the generalization error, and the net effect is that the proposed algorithm has a smaller population risk than the original model. More importantly, the diameter-regularized Hessian-weighted $K$-means algorithm has a better population risk than the original Hessian-weighted $K$-means algorithm.

\subsection{Effect of Diameter Regularization}

\begin{figure}[tb!]
    \centering
    \includegraphics[width=0.45\textwidth]{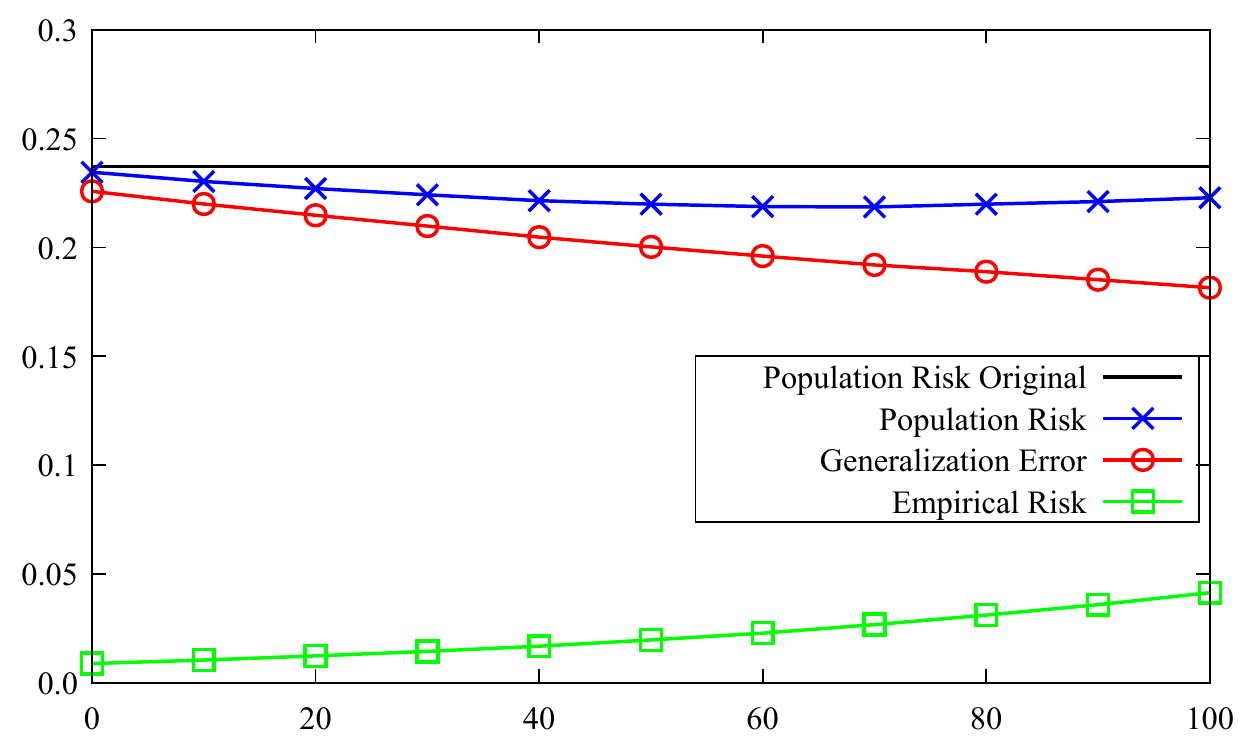}
    \put(-225,25){\rotatebox{90}{Cross Entropy Loss}}
    \put(-110,-10){$\beta$}
    \caption{Diameter-regularized Hessian-weighted $K$-means with different $\beta$ on the MNIST dataset with $K=7$.}
    \label{Fig:reg}
\end{figure}

In Figure \ref{Fig:reg}, we study how $\beta$ affects the performance of our diameter-regularized Hessian-weighted $K$-means algorithm. It can be seen that as $\beta$ increases, the generalization error decreases and distortion in empirical risk increases, which validates the idea that this proposed diameter regularizer can be used to reduce the generalization error. The value of $\beta$ that results in the best population risk can be chosen via cross-validation in practice.

\section{Conclusion}

In this paper, we have provided a theoretical understanding
of how model compression affects the population risk of a
compressed model, which is of practical interest. We
have shown that if the decrease in generalization error due
to model compression can exceed the increase in empirical
risk, the population risk can be improved. Our theoretical
studies convey an important message for designing practical
model compression algorithms. That is to consider the
increase in empirical risk and the decrease in generalization
error jointly, so as to achieve a smaller population risk.

%
%

\bibliography{generalization}
\bibliographystyle{icml2019}

\onecolumn
\appendix
\section{Proof of Lemma \ref{lemma:linear_gen}}
Let $\widetilde{Z}=(\widetilde{X}, \widetilde{Y})$, $\widetilde{X} \in \mathbb{R}^d$ and $\widetilde{Y} \in \mathbb{R}$ denote an independent copy of the training sample $Z_i$. Then, it can be shown that
\begin{align}\label{eq:linear_truth1}
  \mathrm{gen}(\mu,P_{W|S})
   &=\mathbb{E}_{W, S}[L_\mu(W)-L_S(W)] \nn \\
   & =  \mathbb{E}_{W, S}\Big[\mathbb{E}_{\widetilde{Z}}[(\widetilde{Y}-\widetilde{X}^\top W)^2] - \frac{1}{n}\|Y-X^\top W\|_2^2 \Big] \nn \\
   & =  \mathbb{E}_{S}\Big[\mathbb{E}_{\widetilde{Z}}[(\widetilde{Y}-\widetilde{X}^\top(X X^\top)^{-1} X Y)^2] - \frac{1}{n} \|Y-X^\top(X X^\top)^{-1} X Y\|_2^2 \Big],
\end{align}
where $\widetilde{Y} = \widetilde{X}^\top w^*+\widetilde{\varepsilon}$ and $Y = X^\top w^*+\varepsilon$. Then, we have
\begin{align}\label{eq:linear_truth}
   \mathrm{gen}(\mu,P_{W|S})& =  \mathbb{E}_{\varepsilon,\widetilde{\varepsilon},X,\widetilde{X}}
   \big[(\widetilde{\varepsilon}-\widetilde{X}^\top(X X^\top)^{-1} X \varepsilon)^2\big] - \frac{1}{n}\mathbb{E}_{\varepsilon,X}\big[\|\varepsilon-X^\top(X X^\top)^{-1} X \varepsilon\|_2^2\big] \nn \\
   & = \mathbb{E}_{\varepsilon,X,\widetilde{X}}\big[\varepsilon^\top  X^\top(X X^\top)^{-1} \widetilde{X} \widetilde{X}^\top(X X^\top)^{-1}X \varepsilon+ \frac{1}{n} \varepsilon^\top  X^\top(X X^\top)^{-1} {X} \varepsilon\big] \nn \\
   & =  \mathbb{E}_{\varepsilon,X}\big[ {\rm{Tr}}(  X^\top(X X^\top)^{-1}\Sigma_X (X X^\top)^{-1}X \varepsilon \varepsilon^\top) \big]+\frac{\sigma'^2d}{n}\nn\\
   & = \sigma'^2 \mathbb{E}_{X}\big[{\rm{Tr}}(  (X X^\top)^{-1}\Sigma_X ) \big]+\frac{\sigma'^2d}{n}.
\end{align}
Note that $X_i$'s are i.i.d. samples from $\mathcal{N}(0, \Sigma_X)$, then we have $(X X^\top)^{-1}\sim \mathrm{Wishart}^{-1}(\Sigma_X^{-1},n )$, where $\mathrm{Wishart}^{-1}$ denotes the inverse Wishart distribution with $n$ degrees of freedom, and $\mathbb{E}[(X X^\top)^{-1}]=\frac{\Sigma_X^{-1}}{n-d-1}$. It then follows that
\begin{equation}
\mathrm{gen}(\mu,P_{W|S}) = \frac{\sigma'^2} {n-d-1}\big[{\rm{Tr}}(  \Sigma_X^{-1}\Sigma_X ) \big]+\frac{\sigma'^2d}{n}
= \frac{\sigma'^2d}{n}(2+\frac{d+1}{n-d-1}).
\end{equation}

\section{Proof of Theorem \ref{thm:regression}}

For all $\hw\in \hat{\mathcal{W}}$, it can be shown that
\begin{equation}
  \ell(\hw,\widetilde{Z})=(\widetilde{Y}-\widetilde{X}^\top \hw)^2=(\widetilde{X}^\top(w^*-\hw)+\widetilde{\varepsilon})^2.
\end{equation}
Since $\widetilde{X}\sim \mathcal{N}(0, \Sigma_X)$ and $\widetilde{\varepsilon} \sim \mathcal{N}(0,\sigma'^2)$, then $\ell(\hw,\widetilde{Z}) \sim \sigma_\ell^2 \chi^2_1$, where $\sigma_\ell^2 \triangleq (\hw-w^*)^\top \Sigma_X(\hw-w^*)+\sigma'^2 $, and $\chi^2_1$ denotes the chi-squared distribution with one degree of freedom. Then, the CGF of $\ell(\hw,\widetilde{Z})$ is
\begin{equation}
  \Lambda_{\ell(\hw,\widetilde Z)}(\lambda) = -  \sigma_\ell^2 \lambda - \frac{1}{2} \ln(1-2\sigma_\ell^2\lambda), \  \lambda \in (-\infty, \frac{1}{2\sigma_\ell^2}) .
\end{equation}
Thus, $\ell(\hw,\widetilde{Z})$ is not sub-gaussian for all $\lambda \in \mathbb{R}$.
However, It can be shown that
\begin{equation}
  \Lambda_{\ell(\hw,\widetilde Z)}(\lambda) \le \sigma_\ell^4\lambda^2,\quad \lambda<0.
\end{equation}
We need the following lemma from the Theorem 1 of \cite{bu2019tightening} to proceed our analysis.
\begin{lemma}\cite{bu2019tightening}\label{lemma:decouple}
Assume that for all $\hw \in \hat{\mathcal{W}}$, $\Lambda_{\ell(\hw,\widetilde{Z})}(\lambda) \le \frac{\sigma^2 \lambda^2}{2}$ for $\lambda \le 0$. Then,
\begin{align}
    \mathrm{gen}(\mu,P_{\hW|S}) &\le \sqrt{ \frac{2\sigma^2}{n}I(\hW;S)}.
\end{align}
\end{lemma}

Recall  that $C(w^*) = \sup_{\hw \in \hat{\mathcal{W}}}\|\hw-w^*\|_2^2$. We then have the following bound on the CGF of $\ell(\hw,\widetilde Z)$,
\begin{equation}
  \Lambda_{\ell(\hw,\widetilde Z)}(\lambda) \le \lambda^2 \max_{\hw\in \hat{\mathcal{W}}}\sigma_\ell^4 \le \lambda^2 \big(C(w^*)\|\Sigma_X\| +\sigma'^2 \big)^2,\quad \lambda<0.
\end{equation}
Applying Lemma \ref{lemma:decouple} and data processing inequality, we have
\begin{align}
  \mathrm{gen}(\mu,P_{\hW|S})  \le  2\big(C(w^*)\|\Sigma_X\| +\sigma'^2\big)  \sqrt{\frac{I(\hW;W)}{n}}.
\end{align}
\section{Proof of Theorem \ref{thm:DR}}
The constraint on the distortion function can be written as follows:
\begin{align}\label{eq:D_app}
  D & \ge \mathbb{E}_{S,W,\hat{W}}[d_S (\hW,W)] = \frac{1}{n} \mathbb{E}_{S, W,\hat{W}}[(\hW-W)^\top XX^\top (\hW-W)].
\end{align}
It follows from Lemma \ref{lemma:rate_distortion} that
\begin{align}
  R(D) = \min_{P_{\hW|W}}  I(\hW;W), \quad
  \mathrm{s.t.} \quad  \mathbb{E}_{S, W,\hat{W}}[(\hW-W)^\top \frac{1}{n}XX^\top (\hW-W)] \le D.
\end{align}
Note that $\mathbb{E}[W]=w^*$, and $\mathrm{Cov}[W]= \frac{\sigma'^2}{n-d-1} \Sigma_X^{-1}$ since $W$ is the ERM solution. In the following proof, we consider a Gaussian random vector with the same mean and covariance matrix $W_G \sim \mathcal{N}(w^*, \frac{\sigma'^2}{n-d-1} \Sigma_X^{-1})$ as $W$.

For the upper bound of $R(D)$, consider the channel $P^*_{\hW|W} = \mathcal{N}\big((1-\alpha)W+ \alpha w^*, (1-\alpha) \frac{D}{d}\Sigma_X^{-1}\big)$, where $\alpha = \frac{nD}{d\sigma'^2} \le 1$. It can be verified that this channel satisfies the constraint on the distortion:
\begin{align}
  &\mathbb{E}_{S,W,\hat{W}}[d_S (\hW,W)] \nn \\
  &= \alpha^2 \mathbb{E}[(W-w^*)^\top\frac{1}{n}XX^\top(W-w^*)]
  +(1-\alpha)\frac{D}{d}\mathrm{Tr}\Big(\mathbb{E}[\frac{1}{n}XX^\top]\Sigma_X^{-1}\Big) \nn \\
  & = \alpha^2 \mathbb{E}[\big((XX^\top)^{-1}X\varepsilon\big)^\top\frac{1}{n}XX^\top\big((XX^\top)^{-1}X\varepsilon\big)]
  +(1-\alpha)D  \nn \\
  & = \alpha^2\frac{1}{n} \mathbb{E}[\varepsilon^\top X^\top (XX^\top)^{-1}X\varepsilon]
  +(1-\alpha)D \nn\\
  &=D. 
\end{align}
If we let $\xi \sim \mathcal{N}(0,(1-\alpha)\frac{D}{d}\Sigma_X^{-1})$, it follows that
\begin{align}
  R(D) &\le I(W; (1-\alpha)W + \alpha w^* + \xi)  \nn \\
   & \overset{(a)}{\le}  I(W_G; (1-\alpha)W_G+ \xi) \nn \\
   & = \frac{d}{2} \ln\Big( \frac{d\sigma'^2}{(n-d-1)D} -\frac{n}{n-d-1}+1\Big)\nn \\
   & \le \frac{d}{2} \Big(\ln \frac{d\sigma'^2}{(n-d-1)D}\Big)^+,
\end{align}
where (a) is due to the fact that Gaussian distribution maximizes the mutual information in an additive white Gaussian noise channels.

The upper bound of $D(R)$ follows immediately from the upper bound of $R(D)$.

\section{Discussion on Remark \ref{remark:RD}}

Suppose that $\frac{1}{n}XX^\top$ can be approximated by $\Sigma_X$ for large $n$ in  \eqref{eq:D_app}. It then follows that
\begin{align}
  R(D) = \min_{P_{\hW|W}}  I(\hW;W), \quad
  \mathrm{s.t.} \quad  \mathbb{E}_{S, W,\hat{W}}[(\hW-W)^\top \Sigma_X (\hW-W)] \le D.
\end{align}
It can be  easily verified  that the channel $P^*_{W|\hW} = \mathcal{N}(\hW, \frac{D}{d}\Sigma_X^{-1})$ satisfies the distortion constraint.
For any $P_{W|\hW}$ such that $\mathbb{E}_{S,W,\hat{W}}[d_S (\hW,W)] \le D$, it follows that
\begin{align}\label{}
  I(W;\hW) & = \mathbb{E}_{W,\hW} \Big[\ln \frac{P_{W|\hW}}{P_{W}}\Big]\nn \\
   & = \mathbb{E}_{W,\hW} \Big[\ln \frac{P_{W|\hW}}{P^*_{W|\hW}}\Big] + \mathbb{E}_{W,\hW }\Big[\ln \frac{P^*_{W|\hW}}{P_{W_G}}\Big] -\text{KL}(P_W\|P_{W_G}) \nn \\
   & \ge \mathbb{E}_{W,\hW }\Big[\ln \frac{P^*_{W|\hW}}{P_{W_G}}\Big] -\text{KL}(P_W\|P_{W_G}),
\end{align}
where $\text{KL}(P_W\|P_{W_G})$ is the Kullback-Leibler divergence between the two distributions, and the last step follows from the fact that $\text{KL}(P_{W,\hW}\|P^*_{W,\hW})\ge 0$.  Note that
\begin{align}
  &\mathbb{E}_{W,\hW }\Big[\ln \frac{P^*_{W|\hW}}{P_{W_G}}\Big] \nn \\
  &=\frac{d}{2} \ln\frac{d\sigma'^2}{(n-d-1)D}+ \mathbb{E}_{W,\hW }\big[-\frac{d(\hW-W)^\top \Sigma_X(\hW-W)}{2D} + \frac{(n-d-1)(W-w^*)^\top \Sigma_X(W-w^*)}{2\sigma'^2}\big] \nn \\
  & \overset{(a)}{=} \frac{d}{2} \ln\frac{d\sigma'^2}{(n-d-1)D}+ \mathbb{E}_{W,\hW }\big[-\frac{d(\hW-W)^\top \Sigma_X(\hW-W)}{2D} + \frac{d}{2}\big] \nn \\
  & \overset{(b)}{\ge} \frac{d}{2} \ln\frac{d\sigma'^2}{(n-d-1)D},
\end{align}
where (a) follows from the fact that $\mathbb{E}[W]=w^*$ and $\mathrm{Cov}[W]= \frac{\sigma'^2}{n-d-1} \Sigma_X^{-1}$, and (b) is due to the fact that $P_{\hW|W}$ satisfies the distortion constraint. Thus,
\begin{equation}
  R(D) \ge  \frac{d}{2} \ln\frac{d\sigma'^2}{(n-d-1)D}-\text{KL}(P_W\|P_{W_G}).
\end{equation}

\section{Diameter-Regularized Hessian Weighted $K$-means Algorithm}\label{app:algorithm}

We present an iterative algorithm to solve the following diameter-regularized clustering problem.
\begin{eqnarray}
    \min \sum_{k=1}^K \sum_{w^{(j)} \in C^{(k)}} h^{(j)} |w^{(j)} - c^{(k)}|^2 + \beta \max_{k_1, k_2} |c^{(k_1)} - c^{(k_2)}|^2.
\end{eqnarray}

The algorithm alternatively minimizes the objective function over the cluster centroids and assignments.  We first fix centroids, and assign each $w^{(j)}$ to its nearest neighbor. We then fix assignments and update the centroids  by the weighted mean of this cluster. For the farthest pair of centroids, the diameter regularizer pushes them towards each other, so that the output centroids have potentially smaller diameters than those of regular $K$-means.

\begin{algorithm}[htbp]
    \caption{Diameter-regularized Hessian weighted $K$-means}
    \label{algo:diameter_k_means}
\begin{algorithmic}
    \INPUT Weights $\{w^{(1)}, \dots, w^{(d)}\}$, diagonal of Hessian $\{h^{(1)}, \dots, h^{(d)}\}$, diameter regularizer $\beta > 0$, number of clusters $K$, iterations $T$ \\
    \STATE {\bf Initialize} the centroid of $K$ clusters $\{c_0^{(1)}, \dots, c_0^{(K)}\}$
    \FOR{$t=1$ to $T$}
        \STATE {\bf Assignment step:}
            \STATE Initialize $C^{(k)}_t = \emptyset$ for all $k \in [K]$.
            \FOR{$j=1$ to $d$}
                \STATE Assign $w^{(j)}$ to the nearest cluster centroid, i.e. find $k^{(j)}_t = \arg\min_{k \in [K]} |w^{(j)} - c^{(k)}_{t-1}|^2$ and let
                \begin{eqnarray}
                    C_t^{(k^{(j)}_t)} \leftarrow C_t^{(k^{(j)}_t)} \cup \{w^{(j)}\}
                \end{eqnarray}
            \ENDFOR
        \STATE {\bf Update step:}
            \STATE Find current farthest pair of centroids $(k_1, k_2) = \arg\max_{k_1, k_2} |c^{(k_1)}_{t-1} - c^{(k_2)}_{t-1}|^2$.
            \STATE Update $c_t^{(k_1)}$ and $c_t^{(k_2)}$ by
            \begin{eqnarray}
                c_t^{(k_1)} = \frac{\sum_{w^{(j)} \in C_t^{(k_1)}}  h^{(j)} w^{(j)} + \beta c_t^{(k_2)}}{\sum_{w^{(j)} \in C_t^{(k)}}  h^{(j)} + \beta} \,\notag\\
                c_t^{(k_2)} = \frac{\sum_{w^{(j)} \in C_t^{(k_2)}}  h^{(j)} w^{(j)} + \beta c_t^{(k_1)}}{\sum_{w^{(j)} \in C_t^{(k)}}  h^{(j)} + \beta}
            \end{eqnarray}
            \FOR{$k=1$ to $K$, $k \not\in \{k_1, k_2\}$}
                \STATE Update the cluster centroids by
                \begin{eqnarray}
                    c_t^{(k)} = \frac{\sum_{w^{(j)} \in C_t^{(k)}}  h^{(j)} w^{(j)} }{\sum_{w^{(j)} \in C_t^{(k)}}  h^{(j)}}
                \end{eqnarray}
            \ENDFOR
    \ENDFOR
    \OUTPUT Centroids $\{c_T^{(1)}, \dots, c_T^{(K)}\}$ and assignments $\{C_T^{(1)}, \dots, C_T^{(K)}\}$.
\end{algorithmic}
\end{algorithm}

\end{document}